\documentclass[11pt]{article}
\usepackage[hyperref]{acl}
\usepackage{times}
\usepackage{latexsym}
\usepackage{amsmath}
\usepackage{amssymb}
\usepackage{graphicx}
\usepackage{booktabs}
\usepackage{multirow}
\usepackage{xcolor}
\usepackage{pifont}
\usepackage{url}

\newcommand{\Ls}{\lambda_{\mathrm{r}}}

\title{Detecting LLM Hallucinations via Embedding Cluster Geometry:\\A Three-Type Taxonomy with Measurable Signatures}

\author{
  Matic Korun \\
  Independent Researcher \\
  Ljubljana, Slovenia \\
  \texttt{iam.m3x@gmail.com}
}

\begin{document}
\maketitle


\begin{abstract}
We propose a geometric taxonomy of large language model hallucinations based on observable signatures in token embedding cluster structure. By analyzing the static embedding spaces of 11 transformer models spanning encoder (BERT, RoBERTa, ELECTRA, DeBERTa, ALBERT, MiniLM, DistilBERT) and decoder (GPT-2) architectures, we identify three operationally distinct hallucination types: \textbf{Type~1 (center-drift)} under weak context, \textbf{Type~2 (wrong-well convergence)} to locally coherent but contextually incorrect cluster regions, and \textbf{Type~3 (coverage gaps)} where no cluster structure exists. We introduce three measurable geometric statistics: $\alpha$ (polarity coupling), $\beta$ (cluster cohesion), and $\Ls$ (radial information gradient). Across all 11 models, polarity structure ($\alpha > 0.5$) is universal (11/11), cluster cohesion ($\beta > 0$) is universal (11/11), and the radial information gradient is significant (9/11, $p < 0.05$). We demonstrate that the two models failing $\Ls$ significance---ALBERT and MiniLM---do so for architecturally explicable reasons: factorized embedding compression and distillation-induced isotropy, respectively. These findings establish the geometric prerequisites for type-specific hallucination detection and yield testable predictions about architecture-dependent vulnerability profiles.
\end{abstract}


\section{Introduction}

Large language models hallucinate. They generate text that is fluent, confident, and wrong. Despite substantial progress in detecting hallucinations through output-level methods---self-consistency checking \citep{manakul2023selfcheckgpt}, retrieval augmentation, logit calibration \citep{varshney2023stitch}---these approaches share a fundamental limitation: they operate on the model's outputs rather than the geometric structure that produces those outputs. This paper takes a different approach. Rather than asking \textit{whether} a model is hallucinating, we ask \textit{what kind} of geometric failure produced the hallucination, and whether different failure modes leave different measurable signatures in the embedding space.

The core insight is straightforward. Transformer embedding spaces are not homogeneous: they contain cluster structure, where semantically related tokens occupy nearby regions \citep{burns2023discovering}. Different types of generation failures correspond to different geometric relationships between a token's embedding and this cluster structure. A model generating vague, generic content is exhibiting a different geometric pathology than a model confidently committing to the wrong semantic region, which is in turn different from a model encountering a query for which its embedding space has no relevant structure at all.

We formalize this observation as a three-type hallucination taxonomy based on embedding cluster geometry:

\paragraph{Type~1 (Center-drift).} Under weak or ambiguous context, generation collapses toward high-frequency generic tokens near the embedding centroid. The observable signature is low cluster membership and small embedding norm. This produces hallucinations that are plausible but vague---the ``confident generality'' failure mode.

\paragraph{Type~2 (Wrong-well convergence).} The model commits confidently to a locally coherent but contextually inappropriate cluster region. The signature is high local cluster membership with trajectory discontinuities. This produces hallucinations that are detailed, specific, and convincingly wrong---the confabulation failure mode.

\paragraph{Type~3 (Coverage gap).} The query requires semantic combinations absent from training, landing in regions where no cluster structure exists. The signature is weak membership across all clusters and high variance in local similarity. This produces erratic or nonsensical outputs.

Critically, this taxonomy does not require any exotic mathematical machinery to state or validate. It depends on the empirical fact that embedding spaces have cluster structure and that different failure modes produce different geometric signatures in that structure. The detection criteria for each type are operationally distinct and testable with standard tools: $k$-means clustering, cosine similarity, and embedding norms.

To validate the geometric prerequisites for this taxonomy, we introduce three cluster geometry statistics: $\alpha$ (polarity coupling), measuring the strength of antonym-polarity axes within clusters; $\beta$ (cluster cohesion), measuring whether clusters are tighter than background; and $\Ls$ (radial information gradient), quantifying the nonlinear relationship between embedding norm and information content. We compute these statistics across 11 transformer models spanning seven architecture families and two paradigms (encoder and decoder), testing universality across 128- to 1024-dimensional embedding spaces.

Our contributions are: (1)~a three-type hallucination taxonomy grounded in embedding cluster geometry with operationally distinct detection criteria; (2)~three novel embedding-space diagnostics validated across 11 models; (3)~the finding that geometric properties are architecture-independent while their quantitative signatures are architecture-dependent in explicable ways; and (4)~testable predictions linking architectural choices (embedding factorization, distillation) to specific hallucination vulnerability profiles.


\section{Related Work}

\paragraph{Hallucination detection.} The survey by \citet{ji2023survey} and \citet{zhang2023siren} provide comprehensive overviews. Output-level approaches include self-consistency checking \citep{manakul2023selfcheckgpt}, low-confidence generation validation \citep{varshney2023stitch}, and benchmark evaluation \citep{li2023halueval}. Internal-state approaches examine whether models ``know when they're lying'' \citep{azaria2023internal, burns2023discovering}. Our work is complementary: rather than detecting hallucinations post-hoc, we characterize the geometric conditions under which different types arise.

\paragraph{Embedding geometry.} The antonym clustering phenomenon---that words with opposite meanings have high cosine similarity---is well documented \citep{mohammad2013computing, nguyen2016integrating}. Recent work has examined the geometric properties of embedding spaces including isotropy and effective dimensionality. We build on these observations by formalizing cluster-level statistics that serve as diagnostics for hallucination-relevant structure.


\section{Embedding Geometry Preliminaries}

\subsection{Cluster Structure in Token Embeddings}

Transformer models map discrete tokens to continuous vectors via a learned embedding matrix. For a model with vocabulary size $V$ and embedding dimension $d$, this matrix $\mathbf{E} \in \mathbb{R}^{V \times d}$ assigns each token $\tau$ a vector $\mathbf{v}(\tau) \in \mathbb{R}^d$. These embedding spaces exhibit semantic cluster structure: tokens with related meanings occupy nearby regions, forming groups identifiable through standard clustering algorithms.

We extract this structure via $k$-means clustering on the filtered embedding matrix. From each model's vocabulary, we retain only whole-word tokens (excluding subword fragments, special tokens, and tokens without frequency data), yielding between 6{,}322 (ALBERT) and 30{,}932 (GPT-2, RoBERTa) tokens per model. We cluster into $k = 40$ groups using MiniBatchKMeans, producing cluster assignments $c(\tau)$ and centroids $\boldsymbol{\mu}_c$ for each cluster $c$.

\subsection{Polarity Axes and the Antonym Paradox}

A well-known puzzle in distributional semantics is that antonyms---words with opposite meanings---frequently have high cosine similarity and occupy the same embedding clusters \citep{mohammad2013computing, nguyen2016integrating}. This occurs because antonyms share semantic domain (both ``hot'' and ``cold'' relate to temperature) while opposing on a polarity axis. Within each cluster containing antonym pairs, we extract the primary polarity axis via PCA on the difference vectors of co-clustered antonym pairs. This axis represents the direction along which meaning reverses within a shared semantic domain.

\subsection{Self-Information as Radial Diagnostic}

For each token $\tau$ with corpus frequency $f(\tau)$, we compute self-information $I(\tau) = -\log_2 f(\tau)$, measured in bits. This quantity represents the information content of observing token $\tau$: rare words carry high self-information, common words carry low \citep{shannon1948mathematical}. We use self-information rather than the single-term entropy $-f \cdot \log f$ because it is a standard, well-motivated measure of per-token information content and avoids the misleading product that would weight rare words less.


\section{Hallucination Taxonomy}

We classify hallucinations into three types based on the geometric relationship between a generated token's embedding and the ambient cluster structure. Each type has an operationally distinct detection signature.

\subsection{Type~1: Center-Drift}

\paragraph{Mechanism.} When context is weak or ambiguous, the model's hidden state does not receive sufficient signal to navigate to a specific cluster region. Instead, generation drifts toward the centroid of the embedding space---the region of high-frequency, semantically generic tokens.

\paragraph{Observable signature.} (i)~Low soft cluster membership: $H(\mathbf{v})$, defined as the mean top-$k$ cosine similarity to centroids, falls below a threshold. (ii)~Low embedding norm: $\|\mathbf{v}\| < r_{\text{drift}}$, calibrated to the model's norm distribution. (iii)~High token frequency in the generated output. Figure~\ref{fig:type_signatures} (left panel) shows this zone in the norm--membership space.

\subsection{Type~2: Wrong-Well Convergence}

\paragraph{Mechanism.} The model's context signal is strong enough to navigate to a specific cluster region, but the wrong one. The generation locks onto a locally coherent semantic attractor that is contextually inappropriate. This is the most dangerous hallucination type because the model's own confidence metrics will be high.

\paragraph{Observable signature.} (i)~High local cluster membership: $H(\mathbf{v})$ and max centroid similarity are both elevated. (ii)~Trajectory discontinuity: abrupt jumps between cluster regions in a generation sequence. Figure~\ref{fig:type_signatures} (center panel) shows the high-confidence zone.

\subsection{Type~3: Coverage Gap}

\paragraph{Mechanism.} The query requires semantic combinations absent from training. The token's embedding falls in a sparse region where no cluster structure exists.

\begin{figure*}[t]
\centering
\includegraphics[width=\textwidth]{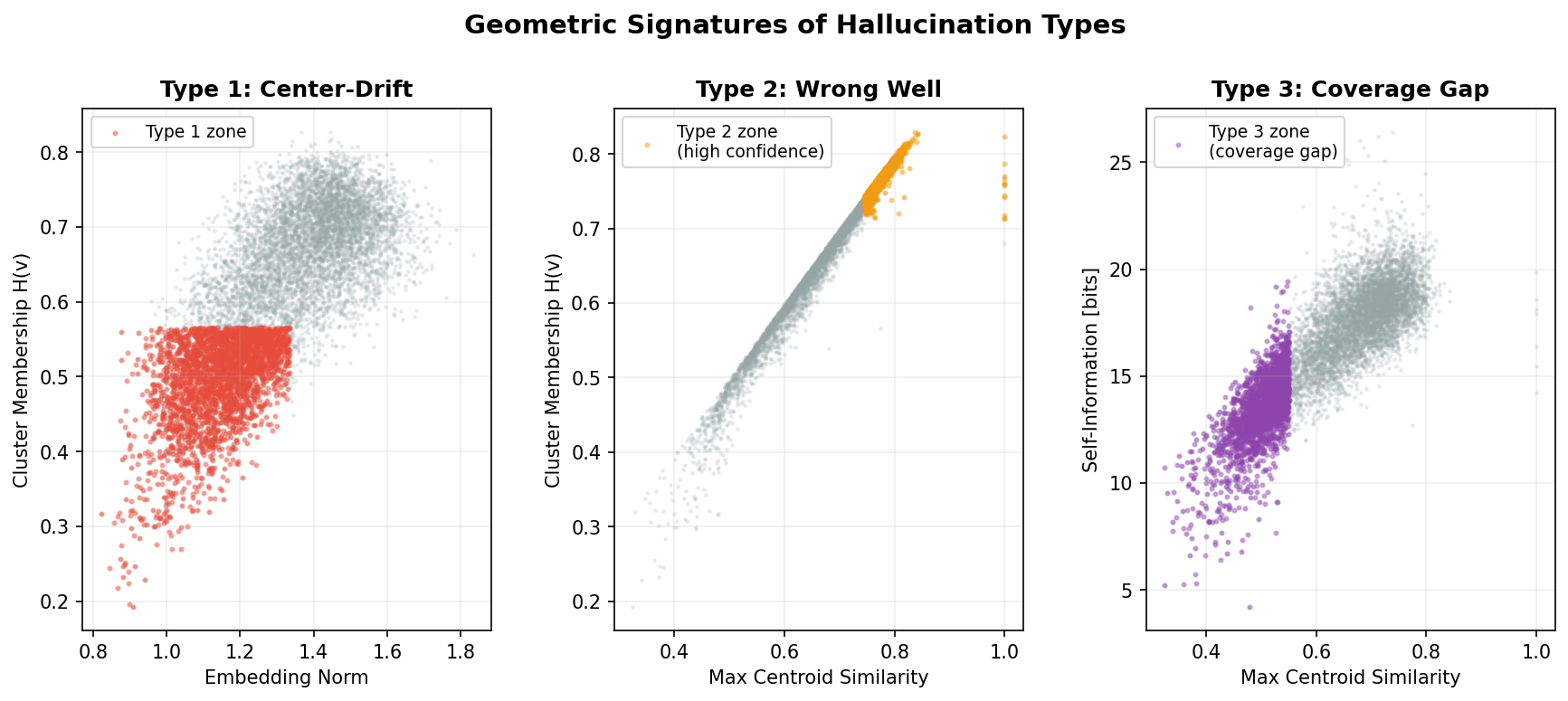}
\caption{Geometric signatures of the three hallucination types in BERT-base embedding space ($k=40$ clusters). \textbf{Left:} Type~1 (center-drift) zone---low norm, low cluster membership. \textbf{Center:} Type~2 (wrong-well) zone---high max centroid similarity, high membership. \textbf{Right:} Type~3 (coverage gap) zone---low max centroid similarity, variable self-information. Colored points indicate tokens falling in each zone; gray points show the full distribution.}
\label{fig:type_signatures}
\end{figure*}

\paragraph{Observable signature.} (i)~Weak membership across all clusters: max centroid similarity is low. (ii)~High variance in local similarity. Figure~\ref{fig:type_signatures} (right panel) shows the coverage gap zone.

\subsection{Geometric Prerequisites}

The taxonomy's validity rests on three measurable conditions: (1)~the embedding space must have real cluster structure ($\beta > 0$); (2)~clusters must have internal organization, including polarity axes ($\alpha > 0$); and (3)~the radial structure must be nonlinear ($\Ls$ significant), indicating that information content varies systematically with distance from the origin. Sections~\ref{sec:lambda} and \ref{sec:cluster_stats} validate these conditions across 11 models.


\section{$\Ls$: Radial Information Gradient}
\label{sec:lambda}

\subsection{Definition}

For each model, we compute the embedding norm $\|\mathbf{v}(\tau)\|$ and self-information $I(\tau)$ for all filtered tokens. We partition the norm range into $B = 40$ equal-width bins (discarding bins with fewer than 10 tokens) and compute the mean self-information per bin. We fit two polynomial models:
\begin{align}
\text{Linear:}\quad \bar{I}(r) &= a_1 r + a_0 \\
\text{Quadratic:}\quad \bar{I}(r) &= \Ls r^2 + a_1 r + a_0
\end{align}
The quadratic coefficient $\Ls$ is our primary diagnostic. We test its significance via nested $F$-test:
\begin{equation}
F = \frac{(\mathrm{SS}_{\mathrm{res,lin}} - \mathrm{SS}_{\mathrm{res,quad}}) / 1}{\mathrm{SS}_{\mathrm{res,quad}} / (n - 3)}
\end{equation}
with corresponding AIC comparison.

\begin{figure}[t]
\centering
\includegraphics[width=\columnwidth]{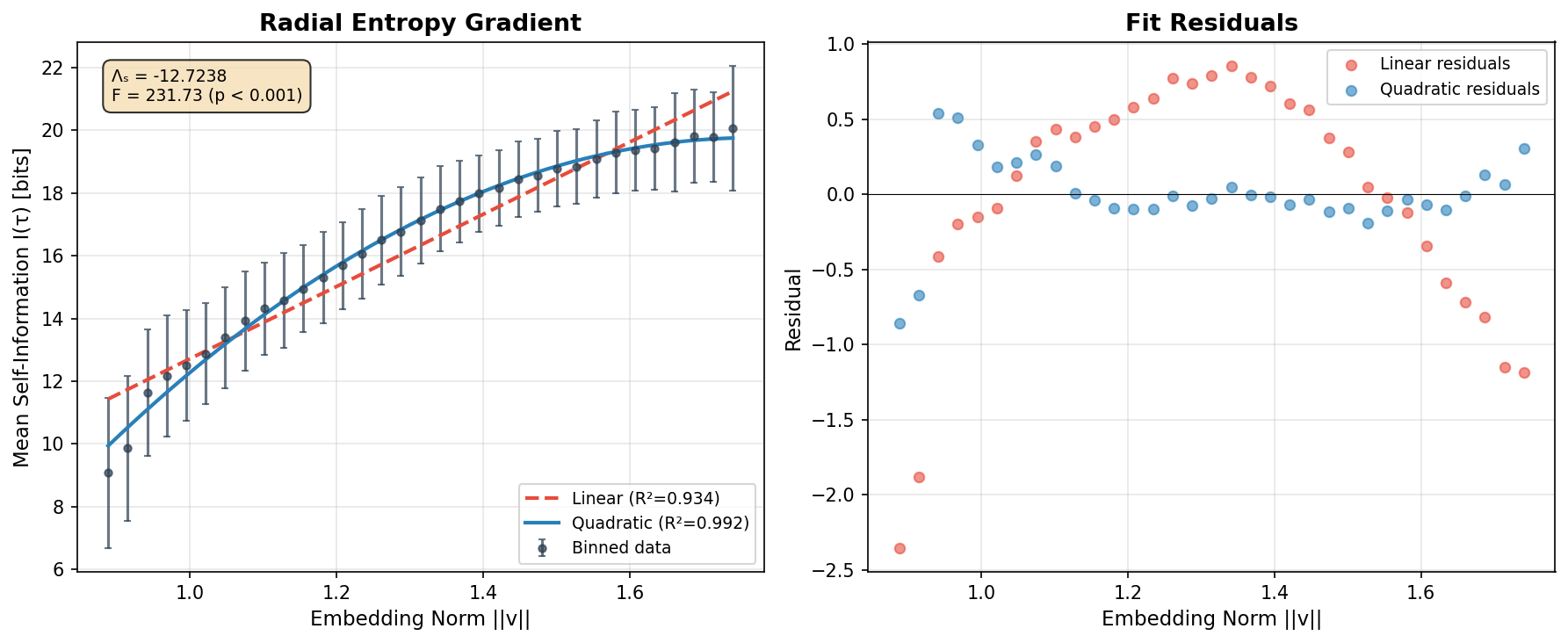}
\caption{Radial information gradient for BERT-base. \textbf{Left:} Mean self-information vs.\ embedding norm with linear and quadratic fits. The quadratic model ($R^2 = 0.992$) significantly outperforms the linear ($R^2 = 0.934$; $F = 231.7$, $p < 0.001$). \textbf{Right:} Residuals show the quadratic fit eliminates the systematic curvature present in linear residuals.}
\label{fig:lambda_s}
\end{figure}

\subsection{Results Across 11 Models}

Table~\ref{tab:main_results} presents the complete results. $\Ls$ is statistically significant ($p < 0.05$) in 9 of 11 models, with $F$-statistics ranging from 10.6 (RoBERTa-large) to 423.0 (ELECTRA-base). The sign of $\Ls$ is predominantly negative (7/11), indicating concave-down radial profiles where self-information increases with norm at a decelerating rate. Two models (DeBERTa, RoBERTa-large) show positive $\Ls$, and two (ALBERT, MiniLM) fail significance.

\begin{table*}[t]
\centering
\small
\begin{tabular}{lrrrrrrrrrr}
\toprule
\textbf{Model} & \textbf{Dim} & \textbf{Tokens} & $\Ls$ & \textbf{$p$-value} & $R^2_{\mathrm{lin}}$ & $R^2_{\mathrm{quad}}$ & $\beta_{\mathrm{diff}}$ & $\alpha$ & $n_\alpha$ & \textbf{Sig.} \\
\midrule
BERT-base       & 768  & 21{,}712 & $-$12.72 & $<$0.001 & 0.934 & 0.992 & 0.157 & 1.02 & 6  & \checkmark \\
BERT-large      & 1024 & 21{,}712 & $-$11.31 & $<$0.001 & 0.922 & 0.989 & 0.234 & 0.91 & 3  & \checkmark \\
DistilBERT      & 768  & 21{,}712 & $-$9.36  & $<$0.001 & 0.871 & 0.980 & 0.111 & 1.07 & 6  & \checkmark \\
RoBERTa-base    & 768  & 30{,}932 & $-$4.05  & $<$0.001 & 0.888 & 0.983 & 0.241 & 0.88 & 5  & \checkmark \\
RoBERTa-large   & 1024 & 30{,}932 & $+$2.47  & 0.003   & 0.063 & 0.359 & 0.214 & 0.87 & 5  & \checkmark \\
ALBERT-base     & 128  & 6{,}322  & $-$28.79 & 0.126   & 0.805 & 0.822 & 0.202 & 0.95 & 5  & \ding{55} \\
ELECTRA-base    & 768  & 21{,}712 & $-$17.95 & $<$0.001 & 0.930 & 0.996 & 0.225 & 0.90 & 9  & \checkmark \\
GPT-2 small     & 768  & 30{,}932 & $-$1.39  & $<$0.001 & 0.953 & 0.984 & 0.212 & 0.91 & 5  & \checkmark \\
GPT-2 medium    & 1024 & 30{,}932 & $-$1.03  & $<$0.001 & 0.977 & 0.990 & 0.216 & 1.01 & 7  & \checkmark \\
DeBERTa-base    & 768  & 30{,}932 & $+$32.15 & $<$0.001 & 0.453 & 0.909 & 0.253 & 1.03 & 8  & \checkmark \\
MiniLM-L6       & 384  & 21{,}712 & $+$3.06  & 0.070   & 0.962 & 0.967 & 0.276 & 1.11 & 11 & \ding{55} \\
\bottomrule
\end{tabular}
\caption{Geometric statistics across 11 transformer models. $\Ls$ = radial information gradient (quadratic coefficient), $\beta_{\mathrm{diff}}$ = cluster cohesion (mean difference in cosine similarity: own centroid minus mean of all other centroids; one-sided $t$-test $p < 0.05$ in all 11 models), $\alpha$ = mean polarity coupling, $n_\alpha$ = number of clusters with $\geq$2 co-clustered antonym pairs. Sig.\ indicates $\Ls$ $p < 0.05$.}
\label{tab:main_results}
\end{table*}

\subsection{Encoder vs.\ Decoder Comparison}

Both GPT-2 variants (decoder-only) show significant negative $\Ls$ with the same qualitative profile as the encoder models: rare words at high norms, common words near the center. The key quantitative difference is magnitude: GPT-2's $\Ls$ values ($-$1.39, $-$1.03) are an order of magnitude smaller than BERT's ($-$12.72), indicating a weaker but still significant radial information gradient. This is consistent with decoder models' training objective (next-token prediction), which distributes information more uniformly across angular dimensions.

\subsection{The Anomalies: ALBERT and MiniLM}
\label{sec:anomalies}

The two non-significant models are architecturally distinctive in ways that directly explain the anomaly.

\paragraph{ALBERT} (128D, $p = 0.126$) uses factorized embeddings: tokens are mapped to a 128-dimensional space before projection to the hidden dimension. Our diagnostic analysis (Table~\ref{tab:anomalies}) reveals that ALBERT utilizes 82\% of its nominal dimensionality (105 of 128 dimensions carry 95\% of variance)---the space is near capacity. The norm coefficient of variation is the lowest of all models (0.105), meaning tokens are compressed into a narrow radial band. The radial gradient exists ($\Ls = -28.79$, the most extreme value) but with high variance: the $R^2$ improvement from linear to quadratic is only 0.072.

\paragraph{MiniLM} (384D, $p = 0.070$) is aggressively distilled. PCA reveals extreme isotropy: the first 100 principal components capture only ${\sim}20\%$ of variance (vs.\ ${\sim}55\%$ for BERT). Mean pairwise cosine similarity is 0.012---essentially zero. Distillation preserved angular structure ($\alpha = 1.11$, $\beta$ significant) while flattening radial structure. The radial profile actually inverts: $\Ls = +3.06$, with rare words at \textit{lower} norms.

\begin{table}[t]
\centering
\scriptsize
\setlength{\tabcolsep}{6pt}
\begin{tabular}{lrrrr}
\toprule
& \textbf{ALBERT} & \textbf{MiniLM} & \textbf{GPT-2} & \textbf{BERT} \\
\midrule
Nominal dim        & 128  & 384  & 768  & 768 \\
Effective dim (95\%) & 105 & 302  & 599  & 635 \\
Utilization        & 0.82 & 0.79 & 0.78 & 0.83 \\
Norm CoV           & 0.105 & 0.142 & 0.088 & 0.112 \\
Mean pairwise cos  & 0.188 & 0.012 & 0.268 & 0.415 \\
$\Ls$              & $-$28.79 & $+$3.06 & $-$1.39 & $-$12.72 \\
$F$ (deg1 vs.\ 2)  & 2.50 & 3.59  & 54.2 & 231.7 \\
$p$-value          & 0.126 & 0.070 & $<$0.001 & $<$0.001 \\
\bottomrule
\end{tabular}
\caption{Deep diagnostic comparison across three anomaly classes plus BERT-base as control. ALBERT shows extreme dimensional compression, MiniLM shows isotropy-induced inversion, and GPT-2 shows decoder-specific radial weakness. All analyses use 40 bins with minimum 10 tokens per bin, consistent with Table~\ref{tab:main_results}.}
\label{tab:anomalies}
\end{table}

\paragraph{Implication.} These anomalies are findings, not weaknesses. Models with compressed embedding spaces (ALBERT) or distillation-induced isotropy (MiniLM) lose radial structure while preserving angular structure. This predicts that such models should be more susceptible to Type~1 (center-drift) hallucinations---a testable hypothesis.

\begin{figure*}[t]
\centering
\includegraphics[width=\textwidth]{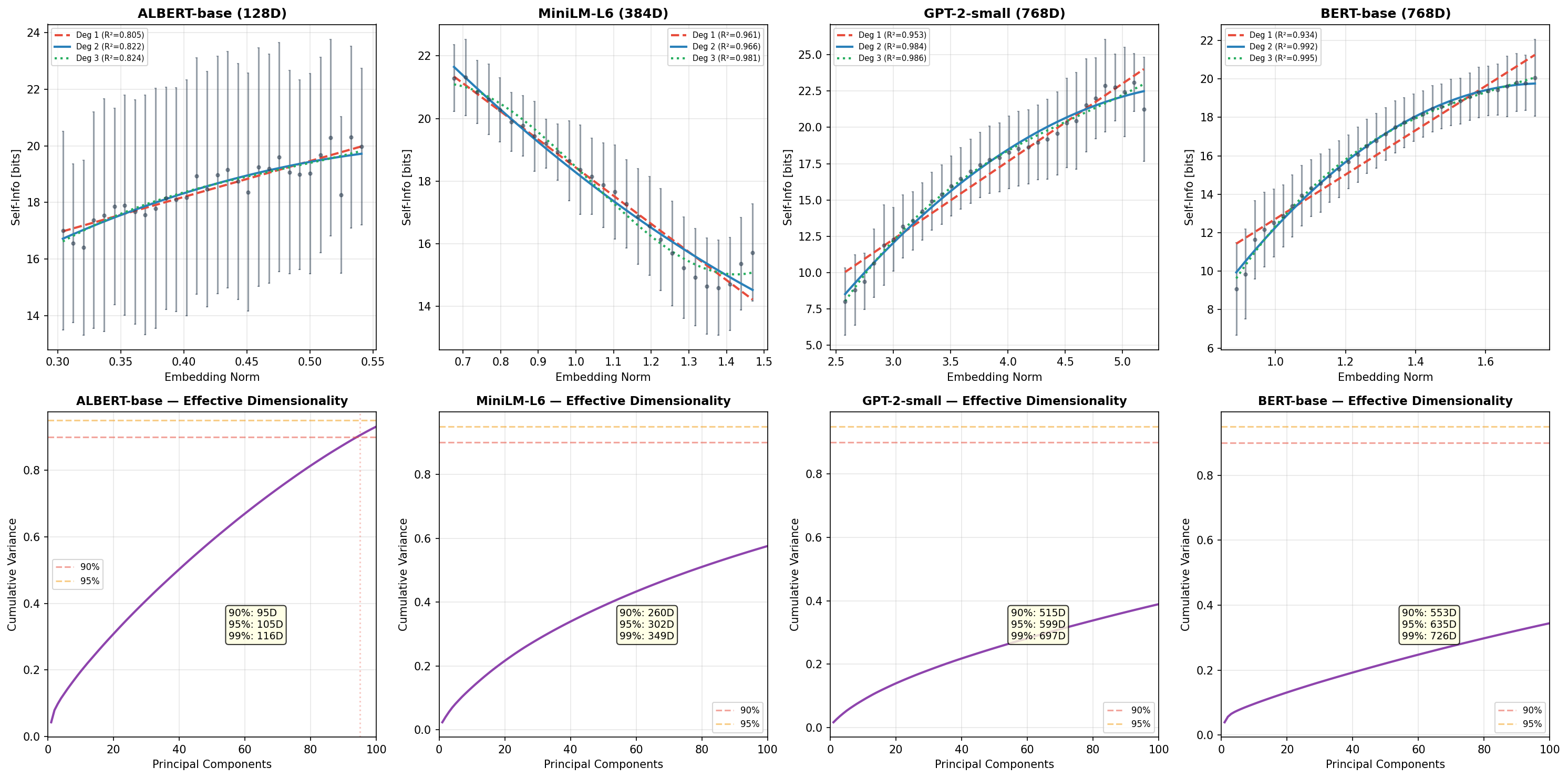}
\caption{Architectural anomaly analysis. \textbf{Top row:} Radial entropy profiles with degree 1--3 polynomial fits. ALBERT (128D) shows high variance from dimensional compression; MiniLM (384D) shows a near-linear inverted profile; GPT-2 (768D) shows an order-of-magnitude weaker gradient than BERT. BERT-base (768D) provides the clear quadratic baseline. \textbf{Bottom row:} Cumulative PCA variance. ALBERT needs nearly all dimensions to capture 95\% variance (space is full). MiniLM distributes variance almost uniformly (extreme isotropy). GPT-2 and BERT show similar utilization profiles despite qualitatively different radial structure.}
\label{fig:anomaly}
\end{figure*}


\section{Cluster Geometry Statistics}
\label{sec:cluster_stats}

\subsection{$\beta$: Cluster Cohesion}

We define cluster cohesion as the mean, over sampled members of cluster $c$, of the difference between similarity to the own centroid and mean similarity to all other centroids:
\begin{equation}
\beta_c = \frac{1}{|S_c|}\sum_{\mathbf{v} \in S_c}\!\Biggl[\mathrm{sim}(\mathbf{v}, \boldsymbol{\mu}_{c}) - \frac{1}{k-1}\sum_{j \neq c} \mathrm{sim}(\mathbf{v}, \boldsymbol{\mu}_j)\Biggr]
\end{equation}
where $S_c$ is a sample of up to 300 members of cluster $c$. A $\beta > 0$ indicates that tokens are closer to their own centroid than to the average of all other centroids---the clusters are real, not artifacts of $k$-means. We evaluate significance via one-sided $t$-test on the per-cluster $\beta_c$ values ($H_0$: mean $\beta \leq 0$). We report this difference formulation ($\beta_{\mathrm{diff}}$) as the primary metric because it is stable across embedding scales; a ratio formulation is sensitive to baseline similarity levels and inflates in near-isotropic spaces.

Results are unambiguous: $\beta$ is significantly positive in all 11 models ($p < 0.001$ in each case). Mean $\beta_{\mathrm{diff}}$ ranges from 0.111 (DistilBERT) to 0.276 (MiniLM), a narrow 2.5$\times$ range consistent with genuine cluster structure across architectures. Across all models, the mean own-centroid similarity (0.50) substantially exceeds the mean other-centroid similarity (0.29).

We note a methodological finding: the \textit{pairwise} $\beta$ variant (within-cluster pairwise similarity vs.\ background) is weaker and frequently non-significant, while the centroid-based $\beta$ is robust. This indicates that clusters are centroid-organized but diffuse---members are reliably closer to their own centroid without being close to each other. This is the expected geometry for clusters containing polarity structure: antonyms share a cluster but are not similar to each other.

\subsection{$\alpha$: Polarity Coupling}

For each cluster containing at least 2 antonym pairs, we extract the primary polarity axis via PCA on antonym difference vectors and compute:
\begin{equation}
\alpha_c = \frac{\text{polarity\_span}}{\text{cluster\_radius}}
\end{equation}
where polarity\_span is the range of projections onto the polarity axis and cluster\_radius is the mean distance from the centroid. An $\alpha$ near or above 1.0 indicates that the polarity axis is a dominant structural feature.

From 231 curated antonym pairs, $\alpha$ exceeds 0.5 in all 11 models, with means ranging from 0.87 (RoBERTa-large) to 1.11 (MiniLM). Notably, DistilBERT---a distilled model---preserves and even strengthens polarity structure relative to its teacher (BERT-base: $\alpha = 1.02$, DistilBERT: $\alpha = 1.07$). Same-cluster antonym pairs show markedly higher cosine similarity (0.65--0.75) compared to cross-cluster pairs (0.40--0.50), confirming the domain-sharing mechanism.

\begin{figure}[t]
\centering
\includegraphics[width=\columnwidth]{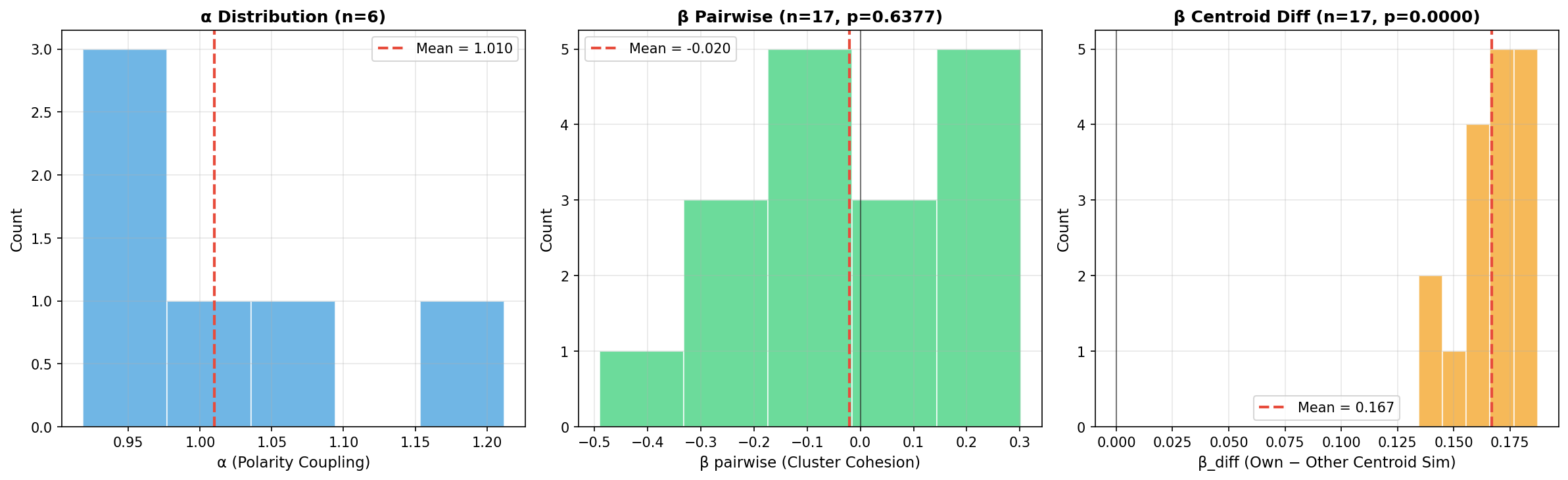}
\caption{Distributions of $\alpha$ (polarity coupling) and $\beta$ (cluster cohesion, both methods) for BERT-base with $k=40$ clusters.}
\label{fig:alpha_beta}
\end{figure}


\section{Detection Architecture}

The three-type taxonomy yields a tiered detection system targeting each hallucination type with increasing computational cost.

\paragraph{Tier~1: Membership Screening (Type~1).} Compute soft cluster membership $H(\mathbf{v}) =$ mean top-5 cosine similarity to centroids. Flag tokens where $H(\mathbf{v}) < \theta_1$ and $\|\mathbf{v}\| < \theta_{\text{norm}}$. Cost: $O(kd)$ per token with precomputed centroids.

\paragraph{Tier~2: Trajectory Analysis (Type~2).} For tokens passing Tier~1, monitor the sequence of cluster assignments across generation. Detect trajectory discontinuities---abrupt jumps between cluster regions indicating contextually inappropriate topic shifts.

\paragraph{Tier~3: Density Analysis (Type~3).} For tokens with low maximum centroid similarity, compute local density via $k$-nearest-neighbor distances. Regions below a density threshold are flagged as coverage gaps.

\paragraph{Architecture-specific calibration.} Our cross-model analysis reveals that detection thresholds must be model-specific. Mean cluster membership $H(\mathbf{v})$ varies from 0.20 to 0.83 across models, and norm distributions vary widely. We recommend percentile-based calibration: Type~1 flagging at the $p_{15}$ percentile of $H(\mathbf{v})$ and $p_{40}$ percentile of norms, adapted per model.


\section{Discussion}

\subsection{What the Framework Measures}

The geometric statistics characterize embedding structure, not truth. A token can have high cluster membership and still be factually wrong if the model has encoded incorrect associations. The taxonomy classifies the geometric \textit{mode of failure}---how the model arrived at its output---not whether the output is correct. This is complementary to, not a replacement for, fact-verification approaches.

\subsection{The $\Ls$ Sign Variation}

The most striking cross-model finding is the sign variation in $\Ls$. Seven models show negative values (concave-down: information saturates at high norms), while DeBERTa and RoBERTa-large show positive values (convex-up: information accelerates at high norms). This suggests that the \textit{direction} of the radial gradient is architecture-dependent, while its \textit{existence} is nearly universal. DeBERTa's disentangled attention mechanism may reorganize the norm-information relationship by separating content and position information. RoBERTa-large shows the weakest radial structure overall ($R^2 = 0.359$ for the quadratic model), suggesting that its embedding space organizes information less radially than other architectures; the significant $F$-test ($p = 0.003$) reflects the large sample size rather than a strong effect.

\begin{figure*}[t]
\centering
\includegraphics[width=\textwidth]{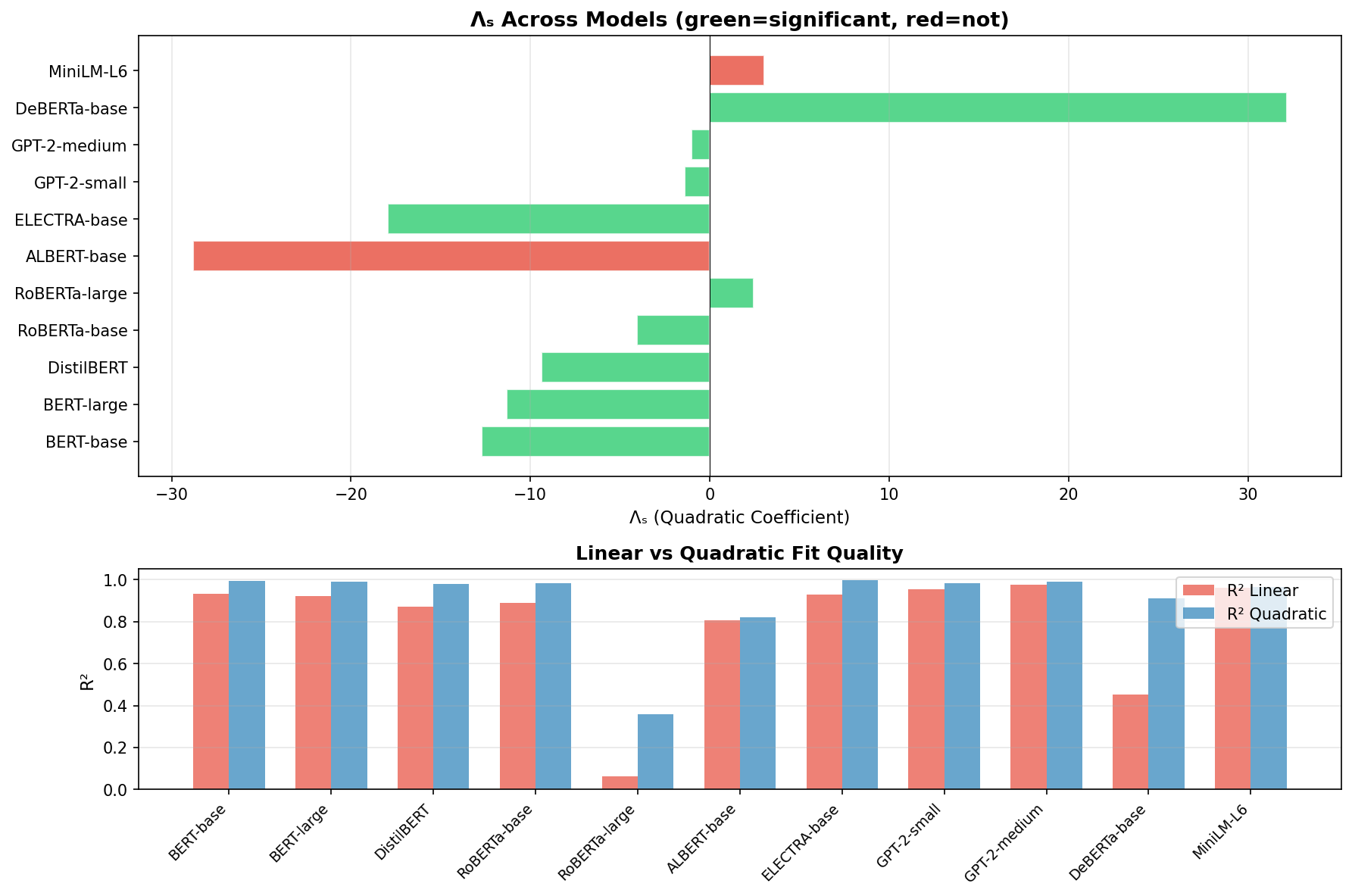}
\caption{$\Ls$ across all 11 models. \textbf{Top:} Magnitude and sign of the quadratic coefficient (green = significant at $p < 0.05$, red = non-significant). \textbf{Bottom:} Linear vs.\ quadratic $R^2$---the quadratic model improves fit in nearly all cases.}
\label{fig:lambda_comparison}
\end{figure*}

\subsection{Compressed Spaces and Vulnerability}

The ALBERT and MiniLM findings generate a testable prediction: models with compressed or isotropic embedding spaces should exhibit higher rates of Type~1 hallucinations, because the radial warning signal is degraded. The angular structure enabling Type~2 detection ($\alpha$, $\beta$) is preserved, but the radial structure distinguishing center-drift from well-placed tokens is not. This creates a detection blind spot specific to these architectures.


\section{Conclusion}

We have proposed and validated a geometric taxonomy of LLM hallucinations, grounded in the empirical cluster structure of token embedding spaces. The three hallucination types---center-drift, wrong-well convergence, and coverage gaps---correspond to distinct geometric signatures measurable with standard tools. The three statistics we introduce ($\alpha$, $\beta$, $\Ls$) are validated across 11 models spanning seven architecture families and both encoder and decoder paradigms.

The key findings are: polarity structure ($\alpha$) is universal across all architectures; cluster cohesion ($\beta$) is universal (11/11 models); and the radial information gradient ($\Ls$) is significant in 9/11 models, with the two exceptions explicable by factorized embeddings and aggressive distillation. The sign variation in $\Ls$ across architectures is itself a finding, suggesting that different architectures organize the norm-information relationship in qualitatively different ways.

The framework yields testable predictions: that compressed-embedding models should show elevated Type~1 vulnerability, and that distilled-isotropic models should show degraded radial detection capability while preserving angular detection. Validating these predictions through controlled hallucination induction, extending to large-scale decoder models, and benchmarking against established evaluation sets are the immediate next steps.


\section*{Limitations}

Several limitations constrain the current work. First, all analysis is on static (input) embeddings, not contextual representations. The cluster structure of static embeddings establishes geometric prerequisites, but hallucination detection in practice operates on contextual hidden states, which may exhibit different geometry. Second, the cross-model survey covers 11 models but no models above 1.5B parameters; extending to larger models requires GPU resources. Third, the detection architecture is proposed but not yet benchmarked against established datasets such as HaluEval \citep{li2023halueval}. Fourth, the $\alpha$ statistic is limited by the number of clusters containing $\geq$2 co-clustered antonym pairs from our curated set; sample sizes range from 3 to 11 clusters per model, and means computed on the smaller samples are less stable. Finally, the $k = 40$ cluster count is a hyperparameter whose sensitivity has not been systematically explored.


\section*{Ethics Statement}

This work analyzes publicly available pretrained language models and uses publicly available word frequency data. No human subjects were involved. The research aims to improve the safety and reliability of language model outputs by enabling more precise hallucination detection. We note that hallucination detection tools could in principle be misused to selectively filter model outputs in ways that introduce bias; we encourage their application toward improving factual accuracy rather than content censorship.


\section*{Acknowledgments}

This work was conducted independently without institutional funding or GPU resources. The author thanks Claude (Anthropic) for assistance with computational pipeline development, statistical validation, and manuscript preparation. All scientific hypotheses, experimental design, and interpretive analysis are the author's own.


\bibliography{references}


\appendix

\section{Reproducibility Protocol}
\label{sec:reproducibility}

All experiments were conducted on an Intel Core i7-6700 CPU (3.40\,GHz, 4 cores, 8 threads) with 16\,GB RAM, running Ubuntu Linux. No GPU was used. The complete pipeline is implemented in Python using PyTorch, Transformers (Hugging Face), scikit-learn, NumPy, SciPy, and matplotlib. Total computation time for the 11-model survey was approximately 90 minutes.

Embedding extraction uses the static word embedding matrix (the input embedding layer) from each model. Vocabulary filtering retains only whole-word tokens with nonzero English frequency via the \texttt{wordfreq} library. Clustering uses MiniBatchKMeans with $k = 40$, batch size 1024, $n_{\text{init}} = 5$, random state 42. The $\beta$ statistic computes the difference between mean cosine similarity to a token's own centroid and its mean similarity to all other centroids; significance is evaluated via one-sided $t$-test on the per-cluster $\beta$ values. The $\alpha$ statistic requires at least 2 co-clustered antonym pairs from a curated set of 231 deduplicated pairs.

Code and data are available at: \url{https://github.com/x3mm3x/llm-hallucination-cluster-geometry}.

\end{document}